\documentclass[10pt,twocolumn,letterpaper]{article}

\usepackage{cvpr}
\usepackage{times}
\usepackage{epsfig}
\usepackage{graphicx}
\usepackage{amsmath}
\usepackage{amssymb}

\usepackage{multirow}
\usepackage{array}
\usepackage{booktabs}
\usepackage{subcaption}
\usepackage{macro}
\usepackage{float}
\newcommand{\nrsfm}{NRS\emph{f}M\xspace}
\newcommand{\sfc}{S\emph{f}C\xspace}
 \renewcommand{\subsubsection}[1]{\vspace{5pt} \noindent \textbf{#1:}}

\usepackage[pagebackref=true,breaklinks=true,letterpaper=true,colorlinks,bookmarks=false]{hyperref}

\cvprfinalcopy 

\def\cvprPaperID{3344} 

\ifcvprfinal\fi
\begin{document}

\title{Deep Interpretable Non-Rigid Structure from Motion}

\author{Chen Kong\\
Carnegie Mellon University\\
{\tt\small chenk@cs.cmu.edu}
\and
Simon Lucey\\
Carnegie Mellon University\\
{\tt\small slucey@cs.cmu.edu}
}

\maketitle

\begin{abstract}
 All current non-rigid structure from motion (NRSfM) algorithms are limited with
 respect to: (i) the number of images, and (ii) the type of shape variability
 they can handle. This has hampered the practical utility of NRSfM for many
 applications within vision. In this paper we propose a novel deep neural
 network to recover camera poses and 3D points solely from an ensemble of 2D
 image coordinates. The proposed neural network is mathematically interpretable
 as a multi-layer block sparse dictionary learning problem, and can handle
 problems of unprecedented scale and shape complexity. Extensive experiments
 demonstrate the impressive performance of our approach where we exhibit
 superior precision and robustness against all available state-of-the-art works.
 The considerable model capacity of our approach affords remarkable generalization to unseen data.
 We propose a quality measure (based on the network weights) which circumvents
 the need for 3D ground-truth to ascertain the confidence we have in the
 reconstruction. Once the network's weights are estimated (for a non-rigid object)
 we show how our approach can effectively recover 3D shape from a single image
 -- outperforming comparable methods that rely on direct 3D supervision.
\end{abstract}

\section{Introduction}
Building an AI capable of inferring the 3D structure and pose of an object from
a single image is a problem of immense importance. Training such a system using
supervised learning requires a large number of labeled images -- how to obtain
these labels is currently an open problem for the vision community.
Rendering~\cite{su2015render} is problematic as the synthetic images seldom
match the appearance and geometry of the objects we encounter in the
real-world. Hand annotation is preferable, but current strategies rely on
associating the natural images with an external 3D dataset (\eg
ShapeNet~\cite{DBLP:journals/corr/ChangFGHHLSSSSX15}, ModelNet~\cite{wu20153d}),
which we refer to as \emph{3D supervision}. If the 3D shape dataset does not
capture the variation we see in the imagery, then the problem is inherently
ill-posed.

\begin{figure}[t]
 \centering
 \begin{subfigure}{\linewidth}
  \centering
  \includegraphics[width=0.24\linewidth]{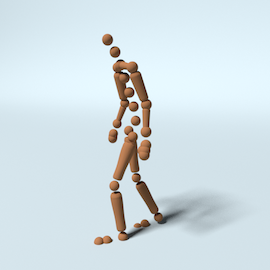}
  \includegraphics[width=0.24\linewidth]{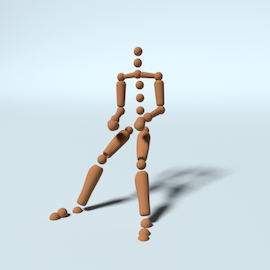}
  \includegraphics[width=0.24\linewidth]{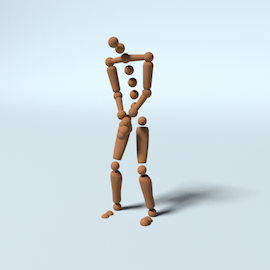}
  \includegraphics[width=0.24\linewidth]{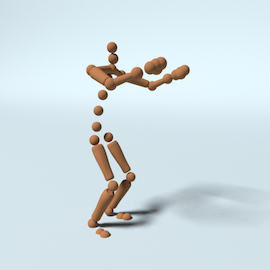}
  \caption{Non-rigid object: moving person.}
 \end{subfigure}

 \begin{subfigure}{\linewidth}
  \centering
  \includegraphics[width=0.24\linewidth]{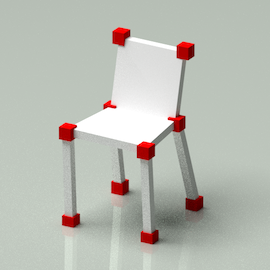}
  \includegraphics[width=0.24\linewidth]{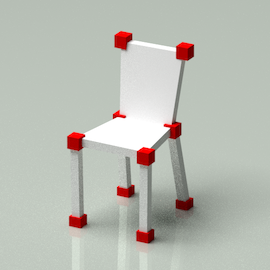}
  \includegraphics[width=0.24\linewidth]{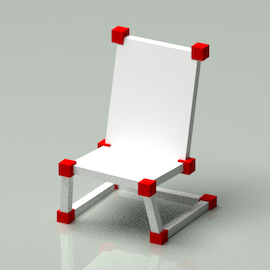}
  \includegraphics[width=0.24\linewidth]{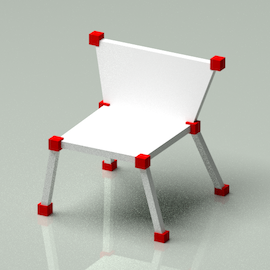}
  \caption{Object category: chair.}
 \end{subfigure}

 \caption{Randomly selected 3D reconstructions by our deep \nrsfm method based
  on multi-layer sparse coding model.
  Spheres in (a) and red cubes in (b) are reconstructed points. Bars, planes
  are manually added for visualization.}
 \label{fig:teaser}
\end{figure}

Non-Rigid Structure from Motion (\nrsfm) offers computer vision a way out of
this quandary -- by recovering the pose and 3D structure of an object category
\emph{solely} from hand annotated 2D landmarks with no need of 3D supervision.
Classically~\cite{bregler2000recovering}, the problem of \nrsfm has been applied
to objects that move non-rigidly over time such as the human body and face. But
\nrsfm is not restricted to non-rigid objects; it can equally be applied to
rigid objects whose object categories deform non-rigidly~\cite{kong2016sfc}.
Consider, for example, the four objects in Figure~\ref{fig:teaser}~(b), our
reconstructions from the visual object category ``chair''. Each object in
isolation represents a rigid chair, but the set of all 3D shapes describing
``chair'' is non-rigid. In other words, each object instance can be modeled as
a deformation from its category's general shape.

Current \nrsfm algorithms~\cite{kumar2018scalable, lee2016consensus,
 chhatkuli2016inextensible} all suffer from the difficulty of processing
large-scale image sequences, limiting their ability to reliably model complex
shape variations. This additionally hinders their ability to generalize to
unseen images. Deep Neural Networks~(DNNs) are an obvious candidate to help
with such issue. However, the influence of DNNs has been most noticeable when
applied to raster representations~(\eg raw pixel
intensities~\cite{deng2009imagenet}). While DNNs have recently exhibited their
success to 3D point representations~(\eg point clouds)~\cite{qi2017pointnet,
 huang2016point}, their use has not been explored in recovering poses and 3D
shapes from an ensemble of vector-based 2D landmarks.

\subsubsection{Contributions}

We propose a novel DNN to solve the problem of \nrsfm. Our employment of DNNs
moves from an opaque black-box to a transparent ``glass-box'' in terms of its
interpretability. The term ``black-box'' is often used as a critique of DNNs
with respect to the general lack of understanding surrounding the inner
workings. We demonstrate how the problem of \nrsfm can be cast as a
multi-layer block sparse dictionary learning problem. Through recent theoretical
innovations~\cite{papyan2017convolutional}, we then show how this problem can be
reinterpreted as a feed-forward DNN auto-encoder that can be efficiently solved
through modern deep learning environments.

Our deep \nrsfm is capable of handling hundreds of thousands of images and
learning large parameterizations to model non-rigidity. Our proposed approach is
completely unsupervised in a 3D sense, relying solely on the projected 2D
landmarks of the non-rigid object or object category to recover the pose and 3D
shape. Our approach dramatically outperforms state-of-the-art methods on a
number of benchmarks, and gets impressive qualitative reconstructions on the
problem of \nrsfm~--~examples of which are shown in Figure~\ref{fig:teaser}.
Moreover, the considerable capacity of modeling non-rigidity allows us to
efficiently apply it to unseen data. This facilitates an accurate 3D
reconstruction of objects from a single view with no aid of 3D ground-truth.
Finally, we propose a measure of model quality~(using coherence and trained
parameters), which improves the practical utility of our model in the real
world applications.

\section{Related Work}

\subsubsection{Non-rigid structure from motion}
\nrsfm is an inherently ill-posed problem since the 3D shapes can vary between
images, resulting in more variables than equations. To alleviate the
ill-posedness, various constraints are exploited including temporal
\footnote{Shapes deform continuously along the sequence of frames.}~\cite{
 akhter2011trajectory, gotardo2011computing, kumar2016multi, kumar2018scalable},
and articulation\footnote{The distance of joints are somehow constant in
 human skeleton.}~\cite{ramakrishna2012reconstructing} priors.
Dai~\etal~\cite{dai2014simple} pioneered the exploration of \nrsfm with minimum
assumptions. They proposed a low-rank model of non-rigidity and a factorization
algorithm recovering both cameras and 3D shapes with no need of additional priors.
The major drawback of this method is the low rank assumption, which highly
restricts the application to complex sequences. To solve this problem, Kong and
Lucey~\cite{kong2016prior} proposed to use an over-complete dictionary with
sparsity to model non-rigid objects and upgraded the factorization algorithm by
characterizing the uniqueness of dictionary learning. However, due to the
enormous parameter space, their method was sensitive to noise and thus had
limited utility in real world applications.

\subsubsection{Structure from category}
\nrsfm has often been criticized as solving a toy problem with few useful
applications beyond being a theoretical curiosity for computer vision. Recently,
Kong~\etal~\cite{kong2016sfc} proposed a novel concept---Structure from
Category~(\sfc)---directly connecting \nrsfm to inferring camera poses and 3D
structure within an ensemble of images stemming from the same object category.
The strength of this approach is the ability to solely use 2D landmarks without
3D supervision.
They provided a convex relaxation solution to this problem. However, the
proposed optimization algorithm could not be applied to large-scale images,
limiting its effectiveness for modeling complex shape variations.

\subsubsection{Single view human pose estimation}
Besides \sfc and \nrsfm, there is another task related to our work, that is
single view human pose estimation. A common solution is assuming that the human
body can be represented through a sparse dictionary.
Ramakrishna~\etal~\cite{ramakrishna2012reconstructing}
proposed to use a matching pursuit algorithm to estimate the sparse
representation. However, since the problem is not convex, their algorithm fails
when initialization is poor. Zhou~\etal~\cite{zhou20153d} proposed to utilize a
convex relaxation to alleviate sensitivities to initialization, but inevitably
introduce additional errors. Another drawback from Zhou~\etal~\cite{zhou20153d,
 ramakrishna2012reconstructing} is its dependence on external 3D models
for estimating the model dictionary (\ie 3D supervision).

\section{Background}
\label{sec: scdnn}
Sparse dictionary learning can be considered as an unsupervised learning task
and divided into two sub-problems: (i) dictionary learning, and (ii)
sparse code recovery. Let us consider sparse code recovery problem, where we
estimate a sparse representation $\zv$ for a measurement vector $\xv$ given the dictionary~$\Wv$
\ie
\begin{equation}
 \min_\zv \Vert \xv - \Wv\zv\Vert_2^2 \quad \st \Vert \zv \Vert_0 < \lambda,
 \label{eq:sparse_coding}
\end{equation}
where $\lambda$ related to the trust region controls the sparsity of recovered
code. One classical algorithm to recover the sparse representation is
Iterative Shrinkage and Thresholding Algorithm
(ISTA)~\cite{daubechies2004iterative, rozell2008sparse,beck2009fast}. ISTA
iteratively executes the following two steps with $\zv^{[0]} = \zero$:
\begin{gather}
 \vv = \zv^{[i]} - \alpha\Wv^T(\Wv\zv^{[i]} - \xv), \\
 \zv^{[i+1]} = \argmin_{\uv} \frac{1}{2}\Vert \uv - \vv \Vert^2_2 + \tau\Vert\uv\Vert_1,
\end{gather}
which first uses the gradient of~$\Vert \xv - \Wv\zv\Vert_2^2$ to
update~$\zv^{[i]}$ in step size $\alpha$ and then finds the closest sparse
solution using an $\ell_1$ convex relaxation. It is well known in literature
that the second step has a closed-form solution using soft thresholding operator.
Therefore, ISTA can be summarized as the following recursive equation:
\begin{equation}
 \zv^{[i+1]} = h_\tau\big(\zv^{[i]} - \alpha\Wv^T(\Wv\zv^{[i]} - \xv)\big),
 \label{eq:ista}
\end{equation}
where $h_\tau$ is a soft thresholding operator and $\tau$ is related to
$\lambda$ for controlling sparsity.

Recently, Papyan~\cite{papyan2017convolutional} proposed to use ISTA and sparse
coding to reinterpret feed-forward neural networks. They argue that feed-forward
passing a single-layer neural network~$\zv = \relu(\Wv^T\xv - b)$ can be
considered as one iteration of ISTA when~$\zv~\ge~0, \alpha=1$ and~$\tau = b$.
Based on this insight, the authors extend this interpretation to feed-forward
neural network with~$n$ layers
\begin{equation}
 \begin{aligned}
  \zv_1 & = \relu(\Wv_1^T\xv - b_1)       \\
  \zv_2 & = \relu(\Wv_2^T\zv_1 - b_2)     \\
        & \quad \vdots                    \\
  \zv_n & = \relu(\Wv_n^T\zv_{n-1} - b_n) \\
 \end{aligned}
\end{equation}
as executing a sequence of single-iteration ISTA, serving as an approximate
solution to the multi-layer sparse coding problem: find~$\{\zv_i\}_{i=1}^n$,
such that
\begin{equation}
 \begin{aligned}
  \xv = \Wv_1\zv_1       & , \quad \Vert \zv_1 \Vert_0 < \lambda_1, \zv_1 \ge 0, \\
  \zv_1 = \Wv_2\zv_2     & , \quad \Vert \zv_2 \Vert_0 < \lambda_2, \zv_2 \ge 0, \\
  \vdots \quad \quad     & , \quad \quad \vdots                                  \\
  \zv_{n-1} = \Wv_n\zv_n & , \quad \Vert \zv_n \Vert_0 < \lambda_n, \zv_n \ge 0, \\
 \end{aligned}
\end{equation}
where the bias terms~$\{b_i\}_{i=1}^n$ (in a similar manner to $\tau$) are
related to~$\{\lambda_i\}_{i=1}^n$, adjusting the sparsity of recovered code.
Furthermore, they reinterpret back-propagating through the deep neural network
as learning the dictionaries~$\{\Wv_i\}_{i=1}^n$. This connection offers a novel
breakthrough for understanding DNNs. In this paper, we extend this to the block
sparse scenario and apply it to solving our \nrsfm problem.

\begin{figure*}[t]
 \centering
 \includegraphics[width=\linewidth]{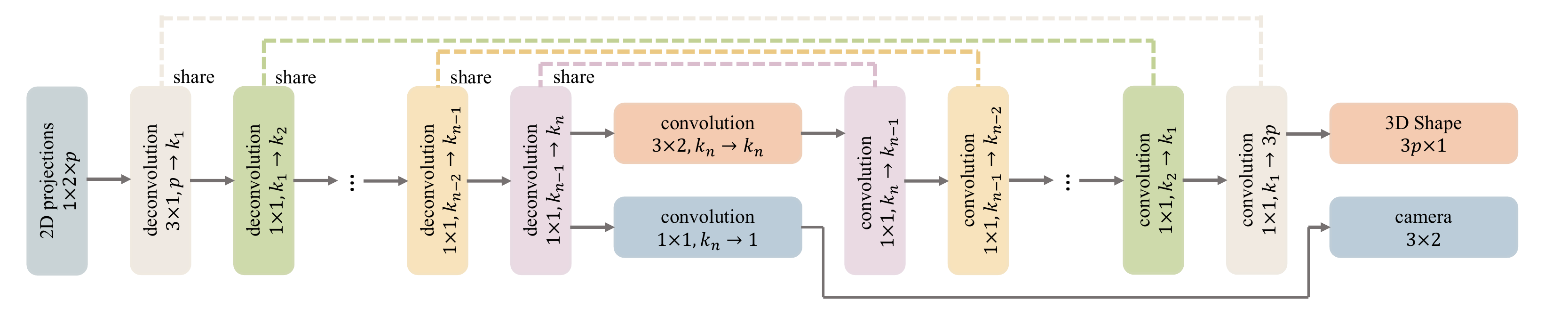}

 \caption{Deep \nrsfm architecture. The network can be divided into two parts:
  encoder and decoder that are symmetric and share convolution kernels (\ie
  dictionaries). The symbol~$a\times b, c \to d$ refers to the operator using
  kernel size~$a\times b$ with~$c$ input channels and $d$~output channels. }
 \label{fig:architecture}
\end{figure*}

\section{Deep Non-Rigid Structure from Motion}
Under weak-perspective projection, \nrsfm deals with the problem of factorizing
a 2D projection matrix $\Wv\in\RR^{p\times 2}$ as the product of a 3D shape
matrix $\Sv\in\RR^{p\times 3}$ and camera matrix $\Mv\in\RR^{3\times 2}$.
Formally,
\begin{equation}
 \Wv = \Sv\Mv,
 \label{eq:proj}
\end{equation}
\begin{equation}
 \Wv = \begin{bmatrix} u_1 & v_1 \\ u_2 & v_2 \\ \vdots & \vdots \\ u_p & v_p \end{bmatrix},~
 \Sv = \begin{bmatrix} x_1 & y_1 & z_1 \\ x_2 & y_2 & z_2 \\ \vdots & \vdots & \vdots \\ x_p & y_p & z_p \end{bmatrix},~
 \Mv^T\Mv = \Iv_2,
\end{equation}
where $(u_i, v_i), (x_i, y_i, z_i)$ are the image and world coordinates of the
$i$-th point. Due to the scale ambiguity between camera focal length and shape
size, we ignore camera scale. The goal of \nrsfm is to recover simultaneously
the shape~$\Sv$ and the camera $\Mv$ for each projection $\Wv$ in a given set
$\mathbb{W}$ of 2D landmarks. In a general \nrsfm including \sfc, this
set~$\mathbb{W}$ could contain deformations of a non-rigid object or various
instances from an object category.

\subsection{Modeling via multi-layer sparse coding}
\label{sec:mlscm}
To alleviate the ill-posedness of \nrsfm and also guarantee sufficient freedom
on shape variation, we propose a novel prior assumption on 3D shapes via
multi-layer sparse coding: The vectorization of $\Sv$ satisfies
\begin{equation}
 \begin{aligned}
  \sv = \Dv_1\psiv_1         & , \quad \Vert \psiv_1 \Vert_0 < \lambda_1, \psiv_1 \ge 0, \\
  \psiv_1 = \Dv_2\psiv_2     & , \quad \Vert \psiv_2 \Vert_0 < \lambda_2, \psiv_2 \ge 0, \\
  \vdots \quad \quad         & , \quad \quad \vdots                                      \\
  \psiv_{n-1} = \Dv_n\psiv_n & , \quad \Vert \psiv_n \Vert_0 < \lambda_n, \psiv_n \ge 0, \\
 \end{aligned}
 \label{eq:mlsc}
\end{equation}
where $\Dv_1 \in \RR^{3p\times k_1}, \Dv_2 \in \RR^{k_1 \times k_2}, \dots,
 \Dv_n \in \RR^{k_{n-1}\times k_n}$ are hierarchical dictionaries.
In this prior, each non-rigid shape is represented by a sequence of hierarchical
dictionaries and corresponding sparse codes. Each sparse code is determined by
its lower-level neighbor and affects the next-level. Clearly this hierarchy
adds more parameters, and thus more freedom into the system. We now show that
it paradoxically results in a more constrained global dictionary and sparse
code recovery.

\subsubsection{More constrained code recovery} In a classical single dictionary
system, the constraint on the representation is element-wise sparsity. Further,
the quality of its recovery entirely depends on the quality of the dictionary.
In our multi-layer sparse coding model, the optimal code not only
minimizes the difference between measurements~$\sv$ and~$\Dv_1\psiv_1$ along
with sparsity regularization~$\Vert \psiv_1 \Vert_0$, but also satisfies
constraints from its subsequent representations. This additional joint
inference imposes more constraints on code recovery, helps to control the
uniqueness and therefore alleviates its heavy dependency on the dictionary
quality.

\subsubsection{More constrained dictionary}
When all equality constraints are satisfied, the multi-layer sparse coding
model degenerates to a single dictionary system. From Equation~\ref{eq:mlsc},
by denoting~$\Dv^{(l)} = \prod_{i=1}^l \Dv_i$, it is implied that
$\sv = \Dv_1\Dv_2\dots\Dv_n\psiv_n = \Dv^{(n)}\psiv_n$.
However, this differs from other single dictionary models~\cite{zhu2014complex,
 zhu2013convolutional, kong2016prior, kong2016sfc, zhou20153d} in terms that a
unique structure is imposed on~$\Dv^{(n)}$~\cite{sulam2017multi}. The
dictionary~$\Dv^{(n)}$ is composed by simpler atoms hierarchically. For
example, each column of $\Dv^{(2)} = \Dv_1\Dv_2$ is a linear combination of
atoms in $\Dv_1$, each column of $\Dv^{(3)} = \Dv^{(2)}\Dv_3$ is a linear
combination of atoms in $\Dv^{(2)}$ and so on. Such a structure results in a
more constrained global dictionary and potentially leads to higher quality with
lower mutual coherence~\cite{donoho2006stable}.

\subsection{Multi-layer block sparse coding}
Given the proposed multi-layer sparse coding model, we now build a conduit
from the proposed shape code~$\{\psiv_i\}_{i=1}^k$ to the 2D projected points.
From Equation~\ref{eq:mlsc}, we reshape vector~$\sv$ to a matrix~$\Sv\in\RR^{p\times3}$ such that $\Sv = \Dv^\sharp_1(\psiv_1 \otimes \Iv_3)$,
where $\otimes$ is Kronecker product and $\Dv_1^\sharp\in\RR^{p\times 3k_1}$ is
a reshape of $\Dv_1$~\cite{dai2014simple}.
From linear algebra, it is well known that~$\Av\Bv \otimes \Iv = (\Av\otimes \Iv)(\Bv\otimes\Iv)$
given three matrices~$\Av, \Bv$, and identity matrix $\Iv$. Based on this lemma, we can derive
that
\begin{equation}
 \small
 \begin{aligned}
  \Sv = \Dv^\sharp_1(\psiv_1 \otimes \Iv_3)                               & , \quad \Vert \psiv_1 \Vert_0 < \lambda_1, \psiv_1 \ge 0, \\
  \psiv_1\otimes \Iv_3 = (\Dv_2 \otimes \Iv_3)(\psiv_2\otimes \Iv_3)      & , \quad \Vert \psiv_2 \Vert_0 < \lambda_2, \psiv_2 \ge 0, \\
  \vdots \quad \quad                                                      & , \quad \quad \vdots                                      \\
  \psiv_{n-1} \otimes \Iv_3 = (\Dv_n \otimes \Iv_3)(\psiv_n\otimes \Iv_3) & , \quad \Vert \psiv_n \Vert_0 < \lambda_n, \psiv_n \ge 0. \\
 \end{aligned}
 \label{eq:mlbsc_s}
\end{equation}

Further, from Equation~\ref{eq:proj}, by right multiplying the camera
matrix~$\Mv\in\RR^{3\times2}$ to the both sides of Equation~\ref{eq:mlbsc_s}
and denote $\Psiv_i = \psiv_i \otimes \Mv$, we obtain that
\begin{equation}
 \small
 \begin{aligned}
  \Wv = \Dv^\sharp_1 \Psiv_1                 & , \quad \Vert \Psiv_1 \Vert_0^{(3\times 2)} < \lambda_1, \\
  \Psiv_1 = (\Dv_2 \otimes \Iv_3)\Psiv_2     & , \quad \Vert \Psiv_2 \Vert_0^{(3\times 2)} < \lambda_2, \\
  \vdots \quad \quad                         & , \quad \quad \vdots                                     \\
  \Psiv_{n-1} = (\Dv_n \otimes \Iv_3)\Psiv_n & , \quad \Vert \Psiv_n \Vert_0^{(3\times 2)} < \lambda_n, \\
 \end{aligned}
 \label{eq:mlbsc_w}
\end{equation}
where $\Vert \cdot \Vert_0^{(3\times 2)}$ divides the argument matrix into
blocks with size $3\times 2$ and counts the number of active blocks.
Since~$\psiv_i$ has active elements less than $\lambda_i$, $\Psiv_i$ has active
blocks less than $\lambda_i$, that is $\Psiv_i$ is block sparse.
This derivation demonstrates that if the shape vector $\sv$ satisfies the
multi-layer sparse coding prior described by Equation~\ref{eq:mlsc}, then its
2D projection~$\Wv$ must be in the format of multi-layer \emph{block} sparse
coding described by Equation~\ref{eq:mlbsc_w}. We hereby interpret \nrsfm
as a hierarchical \emph{block} sparse dictionary learning problem \ie
factorizing $\Wv$ as products of hierarchical dictionaries~$\{\Dv_i\}_{i=1}^n$
and block sparse coefficients~$\{\Psiv_i\}_{i=1}^n$.

\subsection{Block ISTA and DNNs solution}
\label{sec:architecture}

Before solving the multi-layer block sparse coding problem in Equation~\ref{eq:mlbsc_w},
we first consider the single-layer problem:
\begin{equation}
 \min_{\Zv} \Vert \Xv - \Wv\Zv\Vert_F^2 \quad \st~\Vert \Zv \Vert_{0}^{(3\times2)} < \lambda.
\end{equation}
Inspired by ISTA, we propose to solve this problem by iteratively executing
the following two steps:
\begin{gather}
 \Vv = \Zv^{[i]} - \alpha\Wv^T(\Wv\Zv^{[i]} - \Xv), \\
 \Zv^{[i+1]} = \argmin_{\Uv} \frac{1}{2}\Vert \Uv - \Vv \Vert^2_F + \tau\Vert\Uv\Vert_{F1}^{(3\times2)},
\end{gather}
where $\Vert \cdot \Vert_{F1}^{(3\times2)}$ is defined as the summation of
Frobenius norm of each $3\times2$ block, serving as a convex relaxation of
block sparsity constraint. It is derived in~\cite{deng2013group} that the
second step has a closed-form solution computing each block separately by
$\small\Zv^{[i+1]}_j = (h_\tau(\Vert\Vv_j\Vert_F)/\Vert \Vv_j \Vert_F)\Vv_j$,
where the subscript $j$ represents the $j$-th block and $h_\tau$ is a soft
thresholding operator. However, soft thresholding the Frobenius norms for every
block brings unnecessary computational complexity. We show in
the supplementary material that an efficient relaxation is
$\Zv^{[i+1]}_j = h_{b_j}(\Vv_j)$, where $b_j$ is the threshold for the $j$-th
block, controlling its sparsity.
Based on this relaxation, a single-iteration block ISTA with step size
$\alpha=1$ can be represented by :
\begin{equation}
 \Zv = h_{\bv} \big(\Wv^T\Xv\big) = \relu(\Wv^T\Xv - \bv\otimes\one_{3\times 2}),
 \label{eq:singleBISTA}
\end{equation}
where $h_{\bv}$ is a soft thresholding operator using the $j$-th element $b_j$
as threshold of the $j$-th block and the second equality holds if $\Zv$ is
non-negative.

\subsubsection{Encoder}
Recall from Section~\ref{sec: scdnn} that the feed-forward pass through a deep
neural network can be considered as a sequence of single ISTA iterations and
thus provides an approximate recovery of multi-layer sparse codes. We follow
the same scheme: we first relax the multi-layer block sparse coding to be
non-negative and then sequentially use single-iteration block ISTA to solve it \ie
\begin{equation}
 \begin{aligned}
  \Psiv_1 & = \relu((\Dv^\sharp_1)^T\Wv - \bv_1\otimes\one_{3\times 2}),                \\
  \Psiv_2 & = \relu((\Dv_2 \otimes \Iv_3)^T\Psiv_1 - \bv_2\otimes\one_{3\times 2}),     \\
          & \quad \vdots                                                                \\
  \Psiv_n & = \relu((\Dv_n \otimes \Iv_3)^T\Psiv_{n-1} - \bv_n\otimes\one_{3\times 2}), \\
 \end{aligned}
\end{equation}
where thresholds $\bv_1, ..., \bv_n$ are learned, controlling the block sparsity.
This learning is crucial because in previous \nrsfm algorithms utilizing
low-rank~\cite{dai2014simple}, subspaces~\cite{zhu2014complex}
or compressible~\cite{kong2016prior} priors, the weight given to this prior
(\eg rank or sparsity) is hand-selected through a cumbersome cross validation
process. In our approach, this weighting is learned simultaneously with all
other parameters removing the need for any irksome cross validation process.
This formula composes the encoder of our proposed DNN.

\subsubsection{Decoder}
Let us for now assume that we can extract camera $\Mv$ and regular sparse
hidden code $\psiv_n$ from $\Psiv_n$ by some functions \ie $\Mv = \Fc(\Psiv_n)$
and $\psiv_n = \Gc(\Psiv_n)$, which will be discussed in the next section. Then
we can compute the 3D shape vector $\sv$ by:
\begin{equation}
 \begin{aligned}
  \psiv_{n-1} & = \relu(\Dv_n \psiv_n - \bv_n'), \\
              & \quad \vdots                     \\
  \psiv_1     & = \relu(\Dv_2 \psiv_2 - \bv_2'), \\
  \sv         & = \Dv^\sharp_1\psiv_1,
 \end{aligned}
\end{equation}
Note we preserve the ReLU and bias term during decoding to further enforce
sparsity and improve robustness. These portion forms the decoder of our DNN.


\subsubsection{Variation of implementation}
The Kronecker product of identity matrix $\Iv_3$ dramatically increases the
time and space complexity of our approach. To eliminate it and make parameter
sharing easier in modern deep learning environments~(\eg TensorFlow, PyTorch),
we reshape the filters and
features and show that the matrix multiplication in each step of the encoder
and decoder can be equivalently computed via multi-channel $1\times1$
convolution~($*$) and transposed convolution~($*^T$) \ie
\begin{equation}
 (\Dv_1^\sharp)^T\Wv = \dsf_1^\sharp *^T \wsf,
\end{equation}
where {\small $\dsf_1^\sharp \in \RR^{3\times1\times k_1 \times p},
   \wsf\in\RR^{1\times2\times p}$}\footnote{The filter dimension is
 height$\times$width$\times$\# of input channel$\times$\# of output channel.
 The feature dimension is height$\times$width$\times$\# of channel.}.
\begin{equation}
 (\Dv_{i+1}\otimes\Iv_3)^T\Psiv_i = \dsf_{i+1} *^T \Psi_{i},
\end{equation}
where {\small$\dsf_{i+1} \in \RR^{1\times1\times k_{i+1} \times k_i},
   \Psi_i\in\RR^{3\times2\times k_i}.$ }
\begin{equation}
 \Dv_i\psiv_i = \dsf_{i} * \psi_{i},
\end{equation}
where {\small$\dsf_{i} \in \RR^{1\times1\times k_{i} \times k_{i-1}},
   \psi_i\in\RR^{1\times1\times k_i}.$}

\subsubsection{Code and camera recovery}
Estimating~$\psiv_n$ and~$\Mv$ from~$\Psiv_n$ is discussed
in~\cite{kong2016prior} and solved by a closed-form formula. Due to its
differentiability, we could insert the solution directly within our pipeline.
An alternative solution is using a relaxation \ie a fully connected layer
connecting $\Psiv_n$ and $\psiv_n$ and a linear combination among each blocks
of $\Psiv_n$ to estimate~$\Mv$, where the fully connected layer parameters and
combination coefficients are learned from data. In our experiments, we use the
relaxed solution and represent them via convolutions, as shown in
Figure~\ref{fig:architecture}, for conciseness and maintaining proper dimensions.
Since the relaxation has no way to force the orthonormal constraint on the
camera, we seek help from the loss function.

\begin{table*}[]
 \footnotesize
 \centering
 \begin{tabular}{C{0.6cm}|C{1.6cm}C{1.01cm}C{1.01cm}C{1.01cm}C{1.01cm}C{1.01cm}C{1.01cm}C{1.01cm}C{1.01cm}C{1.01cm}C{1.01cm}}
  \specialrule{0.8pt}{1pt}{1pt}
                                                                                         & Subject                             & 1                & 5                & 18              & 23              & 64              & 70              & 102             & 106             & 123             & 127             \\
  \cline{2-12}
                                                                                         & \# of frames                        & 45025            & 13773            & 10024           & 10821           & 11621           & 10788           & 5929            & 12335           & 10788           & 9502            \\
  \hline
                                                                                         & EM-SfM~\cite{torresani2004learning} & 110.23\%         & 119.97\%         & 111.05\%        & 110.94\%        & 114.04\%        & 127.11\%        & 111.60\%        & 113.81\%        & 107.67\%        & 108.07\%        \\
                                                                                         & Simple~\cite{dai2014simple}         & 16.45\%          & 14.07\%          & 13.85\%         & 20.03\%         & 18.13\%         & 18.91\%         & 18.78\%         & 18.63\%         & 19.32\%         & 23.70\%         \\
                                                                                         & Sparse~\cite{kong2016prior}   & 71.23\%          & 66.30\%          & 46.72\%         & 52.44\%         & 70.83\%         & 39.42\%         & 74.12\%         & 47.00\%         & 44.46\%         & 73.85\%         \\
  \parbox[t]{6mm}{\multirow{-4}{*}{\rotatebox[origin=c]{90}{\begin{tabular}{@{}c@{}}Shape \\ Error (\%)\end{tabular}}}} & Ours                                & \textbf{10.74\%} & \textbf{13.40\%} & \textbf{4.73\%} & \textbf{3.24\%} & \textbf{4.38\%} & \textbf{2.17\%} & \textbf{7.32\%} & \textbf{6.83\%} & \textbf{2.23\%} & \textbf{6.00\%} \\
  \hline
                                                                                         & EM-SfM~\cite{torresani2004learning} & 53.1818          & 60.5971          & 53.0413         & 52.2671         & 50.3960         & 56.3713         & 48.5891         & 50.3306         & 47.7355         & 50.8183         \\
                                                                                         & Simple~\cite{dai2014simple}         & 7.9905           & 6.9406           & 6.6340          & 9.5139          & 8.1784          & 8.4294          & 8.0171          & 8.1782          & 8.6922          & 10.9473         \\
                                                                                         & Sparse~\cite{kong2016prior}   & 35.0283          & 35.3014          & 22.6930         & 25.3302         & 32.4681         & 17.7433         & 30.8274         & 21.2735         & 20.3565         & 32.4896         \\
  \parbox[t]{6mm}{\multirow{-4}{*}{\rotatebox[origin=c]{90}{\begin{tabular}{@{}c@{}} Point \\ Error (cm)\end{tabular}}}} & Ours                                & \textbf{5.0638}  & \textbf{6.6717}  & \textbf{2.2664} & \textbf{1.5138} & \textbf{2.2909} & \textbf{0.9622} & \textbf{3.0240} & \textbf{2.9130} & \textbf{0.9844} & \textbf{2.6820} \\
  \specialrule{0.8pt}{1pt}{1pt}
 \end{tabular}
 \caption{Quantitative comparison of our method against the state-of-the-art
  methods in \nrsfm task. Human skeletons are scaled to real-world sizes, around
  1.8 meters high, and the mean point distance is measured in centimeters.}
 \label{tab:nrsfm}
\end{table*}

\subsubsection{Loss function} The loss function must measure the
reprojection error between input 2D points $\Wv$ and reprojected 2D points
$\Sv\Mv$ while simultaneously encouraging orthonormality of the estimated
camera~$\Mv$. One solution is to use spectral norm regularization of~$\Mv$
because spectral norm minimization is the tightest convex relaxation of the
orthonormal constraint~\cite{zhou20153d}. An alternative solution is to hard
code the singular values of~$\Mv$ to be exact ones with the help of Singular
Value Decomposition~(SVD). Even though SVD is generally non-differentiable, the
numeric computation of SVD is differentiable and most deep learning packages
implement its gradients~(\eg PyTorch, TensorFlow). In our implementation and
experiments, we use SVD to ensure the success of the orthonormal constraint and
a simple Frobenius norm to measure reprojection error,
\begin{equation}
 Loss = \Vert \Wv - \Sv\tilde{\Mv} \Vert_F, \quad \tilde{\Mv} = \Uv\Vv^T,
\end{equation}
where $\Uv\Sigmav\Vv^T = \Mv$ is the SVD of the camera matrix.

\section{Experiments}
We conduct extensive experiments to evaluate the performance of our deep
solution for solving \nrsfm and \sfc problems. Further, for evaluating
generalizability, we conduct an experiment applying the pre-trained DNN
to unseen data and reconstruct 3D human pose from a single view. Note that
in all experiments, our model has no access to 3D ground-truth except
qualitative and quantitative evaluations for comparison against the
state-of-art methods. A detailed description of our architectures is in the
supplementary material.

\subsection{NRS\textbf{\textit{f}}M on CMU Motion Capture}
\begin{figure*}[t]
 \centering
 \begin{subfigure}{0.3\textwidth}
  \includegraphics[width=0.32\linewidth]{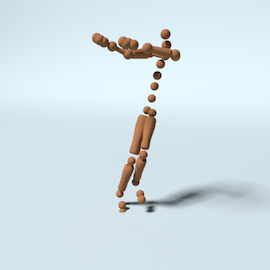}
  \includegraphics[width=0.32\linewidth]{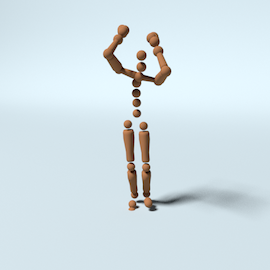}
  \includegraphics[width=0.32\linewidth]{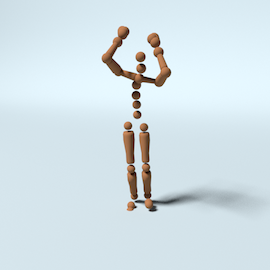}
  \caption{Subject 01.}
 \end{subfigure}
 \hspace{4mm}
 \begin{subfigure}{0.3\textwidth}
  \includegraphics[width=0.32\linewidth]{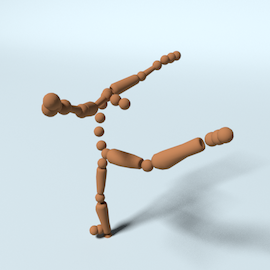}
  \includegraphics[width=0.32\linewidth]{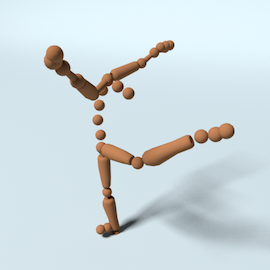}
  \includegraphics[width=0.32\linewidth]{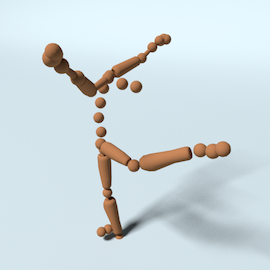}
  \caption{Subject 05.}
 \end{subfigure}
 \hspace{4mm}
 \begin{subfigure}{0.3\textwidth}
  \includegraphics[width=0.32\linewidth]{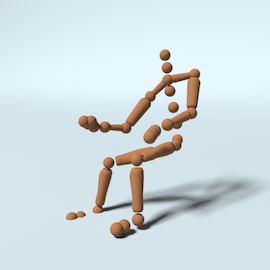}
  \includegraphics[width=0.32\linewidth]{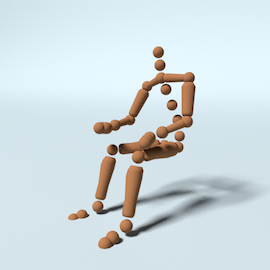}
  \includegraphics[width=0.32\linewidth]{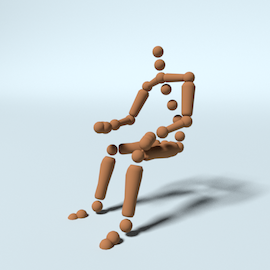}
  \caption{Subject 18.}
 \end{subfigure}

 \begin{subfigure}{0.3\textwidth}
  \includegraphics[width=0.32\linewidth]{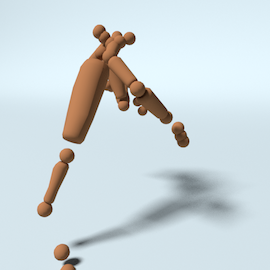}
  \includegraphics[width=0.32\linewidth]{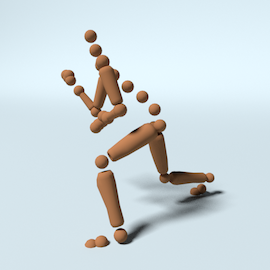}
  \includegraphics[width=0.32\linewidth]{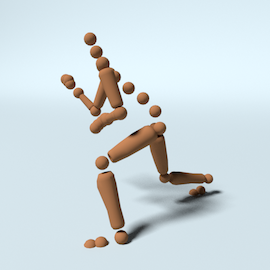}
  \caption{Subject 23.}
 \end{subfigure}
 \hspace{4mm}
 \begin{subfigure}{0.3\textwidth}
  \includegraphics[width=0.32\linewidth]{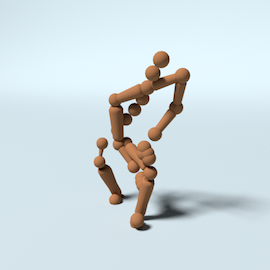}
  \includegraphics[width=0.32\linewidth]{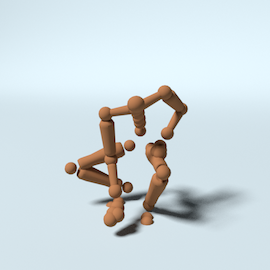}
  \includegraphics[width=0.32\linewidth]{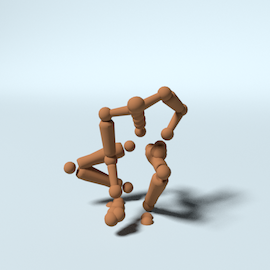}
  \caption{Subject 48.}
 \end{subfigure}
 \hspace{4mm}
 \begin{subfigure}{0.3\textwidth}
  \includegraphics[width=0.32\linewidth]{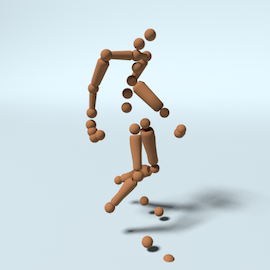}
  \includegraphics[width=0.32\linewidth]{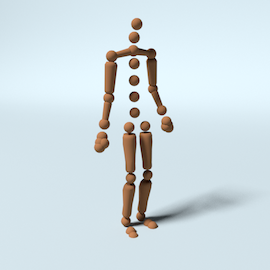}
  \includegraphics[width=0.32\linewidth]{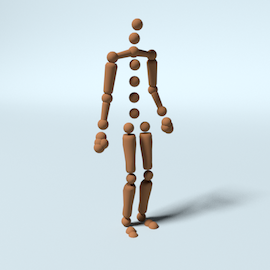}
  \caption{Subject 70.}
 \end{subfigure}

 \begin{subfigure}{0.3\textwidth}
  \includegraphics[width=0.32\linewidth]{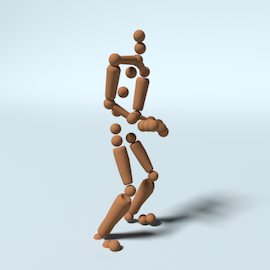}
  \includegraphics[width=0.32\linewidth]{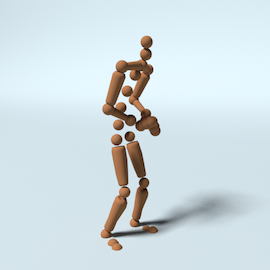}
  \includegraphics[width=0.32\linewidth]{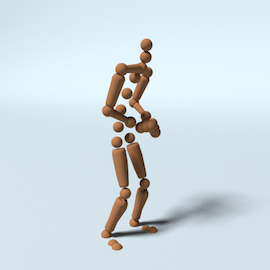}
  \caption{Subject 102.}
 \end{subfigure}
 \hspace{4mm}
 \begin{subfigure}{0.3\textwidth}
  \includegraphics[width=0.32\linewidth]{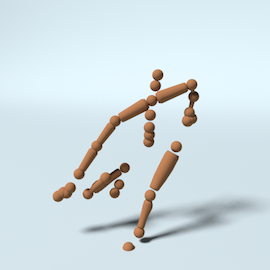}
  \includegraphics[width=0.32\linewidth]{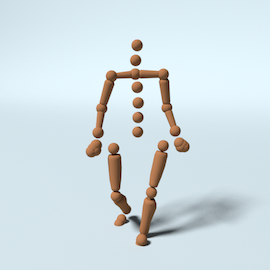}
  \includegraphics[width=0.32\linewidth]{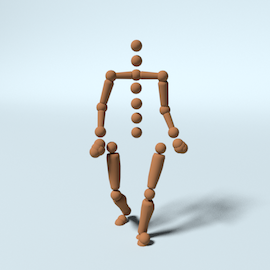}
  \caption{Subject 106.}
 \end{subfigure}
 \hspace{4mm}
 \begin{subfigure}{0.3\textwidth}
  \includegraphics[width=0.32\linewidth]{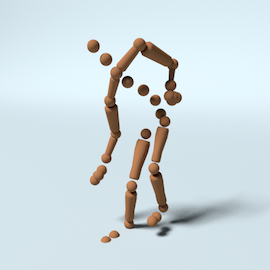}
  \includegraphics[width=0.32\linewidth]{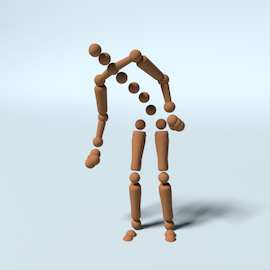}
  \includegraphics[width=0.32\linewidth]{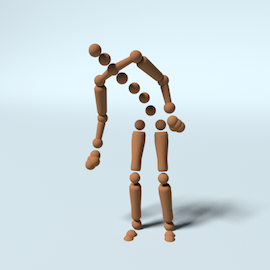}
  \caption{Subject 123.}
 \end{subfigure}

 \begin{subfigure}{0.3\textwidth}
  \includegraphics[width=0.32\linewidth]{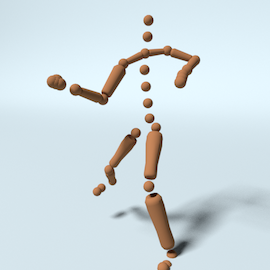}
  \includegraphics[width=0.32\linewidth]{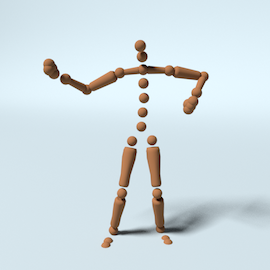}
  \includegraphics[width=0.32\linewidth]{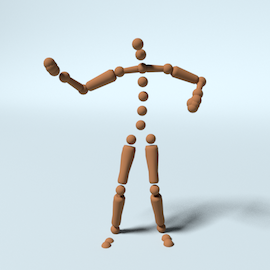}
  \caption{Subject 127.}
 \end{subfigure}
 \hspace{4mm}
 \begin{subfigure}{0.3\textwidth}
  \includegraphics[width=0.32\linewidth]{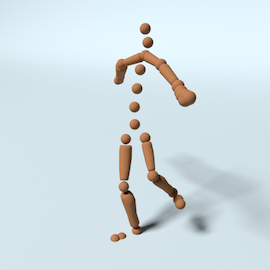}
  \includegraphics[width=0.32\linewidth]{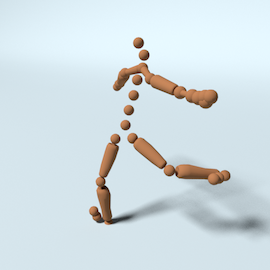}
  \includegraphics[width=0.32\linewidth]{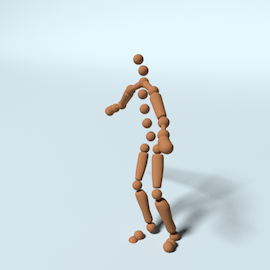}
  \caption{Failure case.}
 \end{subfigure}
 \hspace{4mm}
 \begin{subfigure}{0.3\textwidth}
  \includegraphics[width=0.32\linewidth]{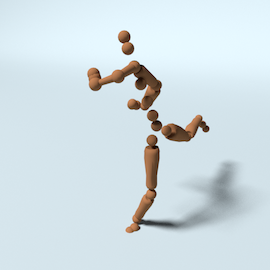}
  \includegraphics[width=0.32\linewidth]{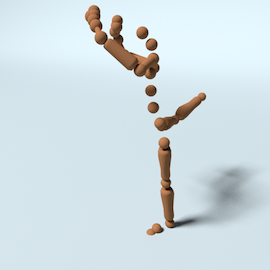}
  \includegraphics[width=0.32\linewidth]{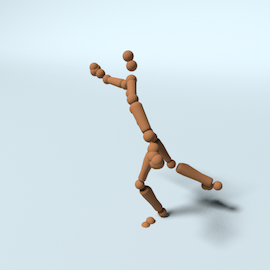}
  \caption{Failure case.}
 \end{subfigure}

 \caption{Qualitative evaluation on reconstructed human skeletons. (a) to (j)
  are randomly selected from each subject. (k) and (l) are two failure cases
  with the largest errors among all 140,606 images. In each sub-figure, the
  left is the reconstruction of~\cite{dai2014simple}, the middle is the
  ground-truth, and the right is ours.}
 \label{fig:nrsfm}

 \vspace{5mm}

 \begin{subfigure}{\textwidth}
  \centering
  \includegraphics[height=1.8cm]{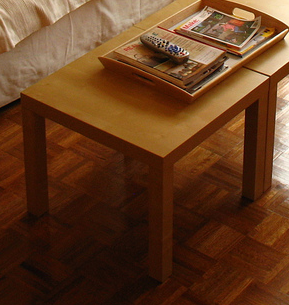}
  \includegraphics[height=1.8cm]{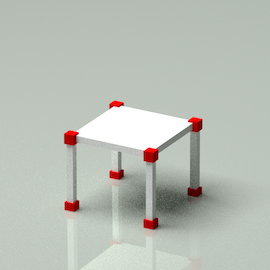}
  \includegraphics[height=1.8cm]{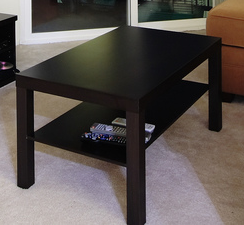}
  \includegraphics[height=1.8cm]{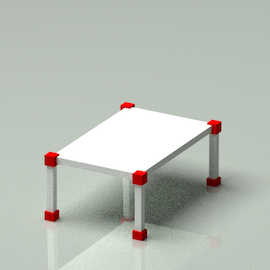}
  \includegraphics[height=1.8cm]{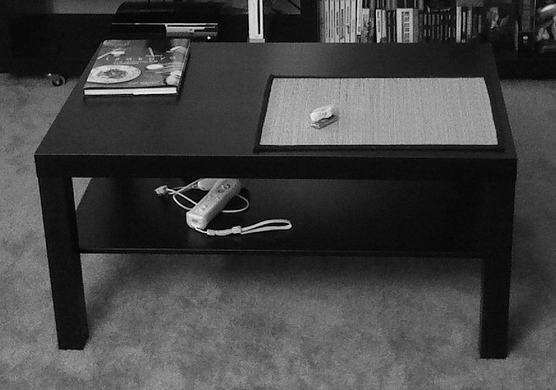}
  \includegraphics[height=1.8cm]{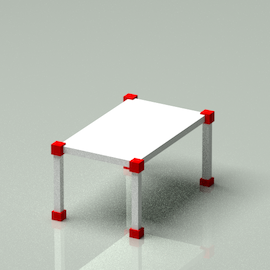}
  \includegraphics[height=1.8cm]{figures/table_096_img.png}
  \includegraphics[height=1.8cm]{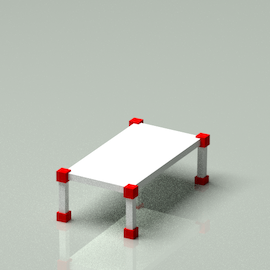}
  \caption{Object table.}
 \end{subfigure}

 \begin{subfigure}{\textwidth}
  \centering
  \includegraphics[height=1.8cm]{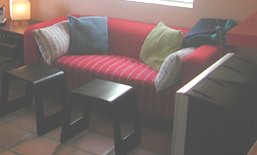}
  \includegraphics[height=1.8cm]{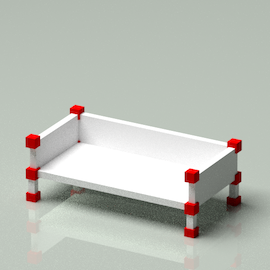}
  \hspace{0.7mm}
  \includegraphics[height=1.8cm]{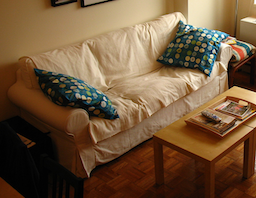}
  \includegraphics[height=1.8cm]{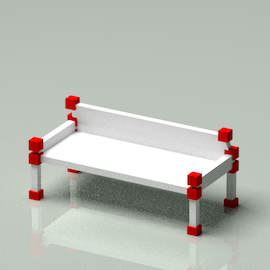}
  \hspace{0.7mm}
  \includegraphics[height=1.8cm]{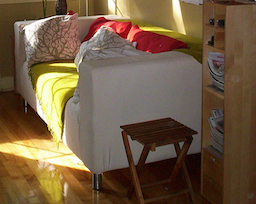}
  \includegraphics[height=1.8cm]{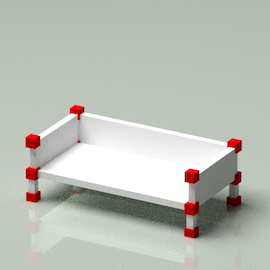}
  \hspace{0.7mm}
  \includegraphics[height=1.8cm]{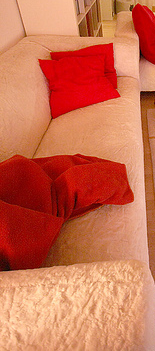}
  \includegraphics[height=1.8cm]{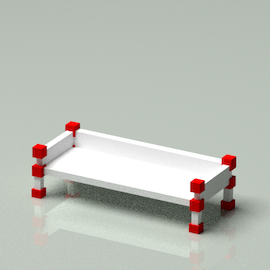}
  \caption{Object sofa.}
 \end{subfigure}

 \begin{subfigure}{\textwidth}
  \centering
  \includegraphics[height=1.8cm]{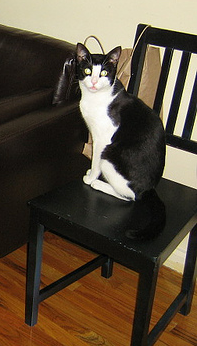}
  \includegraphics[height=1.8cm]{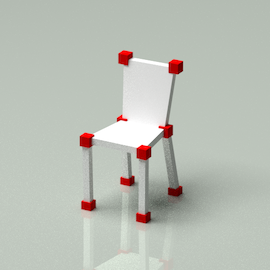}
  \hspace{0.7mm}
  \includegraphics[height=1.8cm]{figures/chair_125_img.png}
  \includegraphics[height=1.8cm]{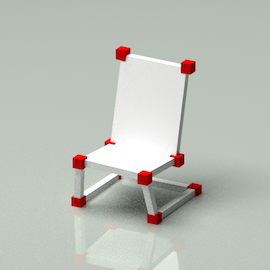}
  \hspace{0.7mm}
  \includegraphics[height=1.8cm]{figures/chair_002_img.png}
  \includegraphics[height=1.8cm]{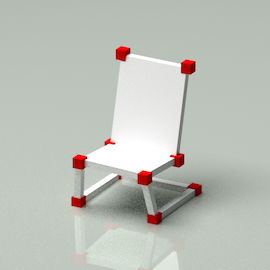}
  \hspace{0.7mm}
  \includegraphics[height=1.8cm]{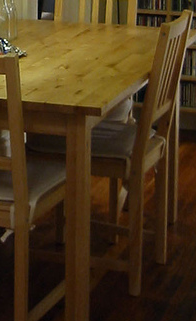}
  \includegraphics[height=1.8cm]{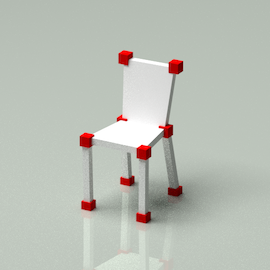}
  \hspace{0.7mm}
  \includegraphics[height=1.8cm]{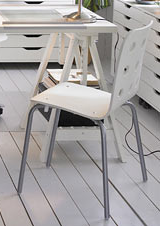}
  \includegraphics[height=1.8cm]{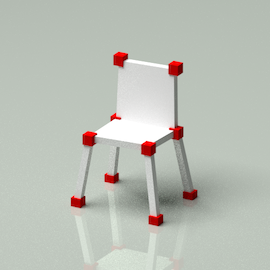}
  \caption{Object chair.}
 \end{subfigure}

 \begin{subfigure}{\textwidth}
  \centering
  \includegraphics[height=1.8cm]{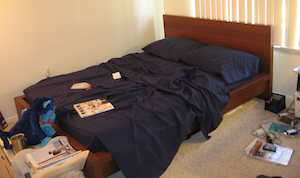}
  \includegraphics[height=1.8cm]{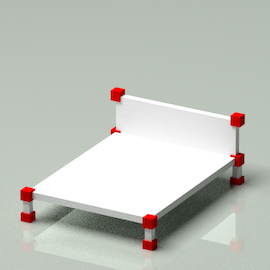}
  \hspace{3mm}
  \includegraphics[height=1.8cm]{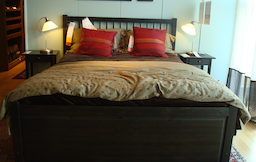}
  \includegraphics[height=1.8cm]{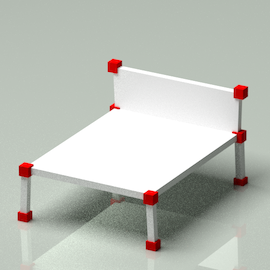}
  \hspace{3mm}
  \includegraphics[height=1.8cm]{figures/bed_043_img.png}
  \includegraphics[height=1.8cm]{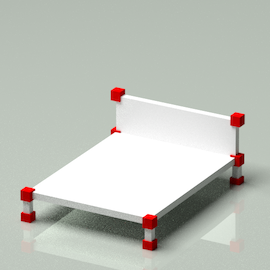}
  \caption{Object bed.}
 \end{subfigure}

 \caption{Qualitative results of \sfc task. Reconstructions are randomly
 selected from each object category. Red cubes are reconstructed points while
 the planes and bars are manually added for descent rendering.}
 \label{fig:sfc}
\end{figure*}

We first apply our method to solving the problem of \nrsfm using the CMU motion
capture  dataset\footnote{http://mocap.cs.cmu.edu/}. For evaluation on complex
sequences, we concatenate all motions of the same subject and select ten
subjects from CMU MoCap so that each subject contains tens of thousands of
frames. We randomly create orthonormal cameras for each frame to project the 3D
human joints onto images. We compare our method against state-of-the-art
\nrsfm works with code released online\footnote{
 Paladini~\etal~\cite{paladini2009factorization} fails on all sequences and
 therefore removed from the table. Works~\cite{taylor2010non, del2007non,
  vicente2012soft, hamsici2012learning, chhatkuli2016inextensible,
  gotardo2011kernel, lee2016consensus, gotardo2011non} did not release code.
 Works~\cite{akhter2011trajectory, gotardo2011computing, kumar2018scalable,
  kumar2016multi} use additional priors, say temporal continuity, and thus not
 applicable.}~\cite{torresani2004learning, dai2014simple, kong2016prior}.
Since none of them are capable of scaling up to this number of frames, we
shuffle each sequence, divide them into mini batches
each containing 500 frames, feed each mini batch into baselines, and then
compute the mean error. Our model is trained on the entire sequence. For error
metrics, we use the shape error ratio defined as
$\small\frac{1}{\vert \Sc \vert}\sum_{\Sc} \frac{\Vert \Sv - \hat{\Sv} \Vert_F}{\Vert \hat{\Sv} \Vert_F},$
where $\hat{\Sv}$ is the 3D ground-truth and $\Sc$ is the set of all shapes;
as well as the mean point distance defined as
$\frac{1}{\vert \Sc \vert}\sum_{\Sc} \sum_i \frac{\Vert\Sv_i - \hat{\Sv}_i \Vert_2}{p},$
where $\Sv_i$ is 3D coordinates of $i$-th point on shape $\Sv$ and $p$ is the
number of points. Note that shapes are normalized to real-world sizes so that
each human skeleton is around 1.8 meters high, and the mean point distance is
computed in centimeters. The results are summarized in Table~\ref{tab:nrsfm}.
One can see that our method obtains impressive reconstruction performance and
outperforms others in every sequences. We randomly select a frame for each
subject and render the reconstructed human skeleton in
Figure~\ref{fig:nrsfm}~(a) to \ref{fig:nrsfm}~(j). To give a sense of the quality of
reconstructions when our method fails, we go through all ten subjects in a total
of 140,606 frames and select the frames with the largest errors as shown in
Figure~\ref{fig:nrsfm}(k) and \ref{fig:nrsfm}~(l). Even in the worst cases, our
method grasps a rough 3D geometry of human body instead of completely diverging.

\subsubsection{Noise performance} 
\begin{figure}[t]
 \centering
 \includegraphics[width=0.9\linewidth]{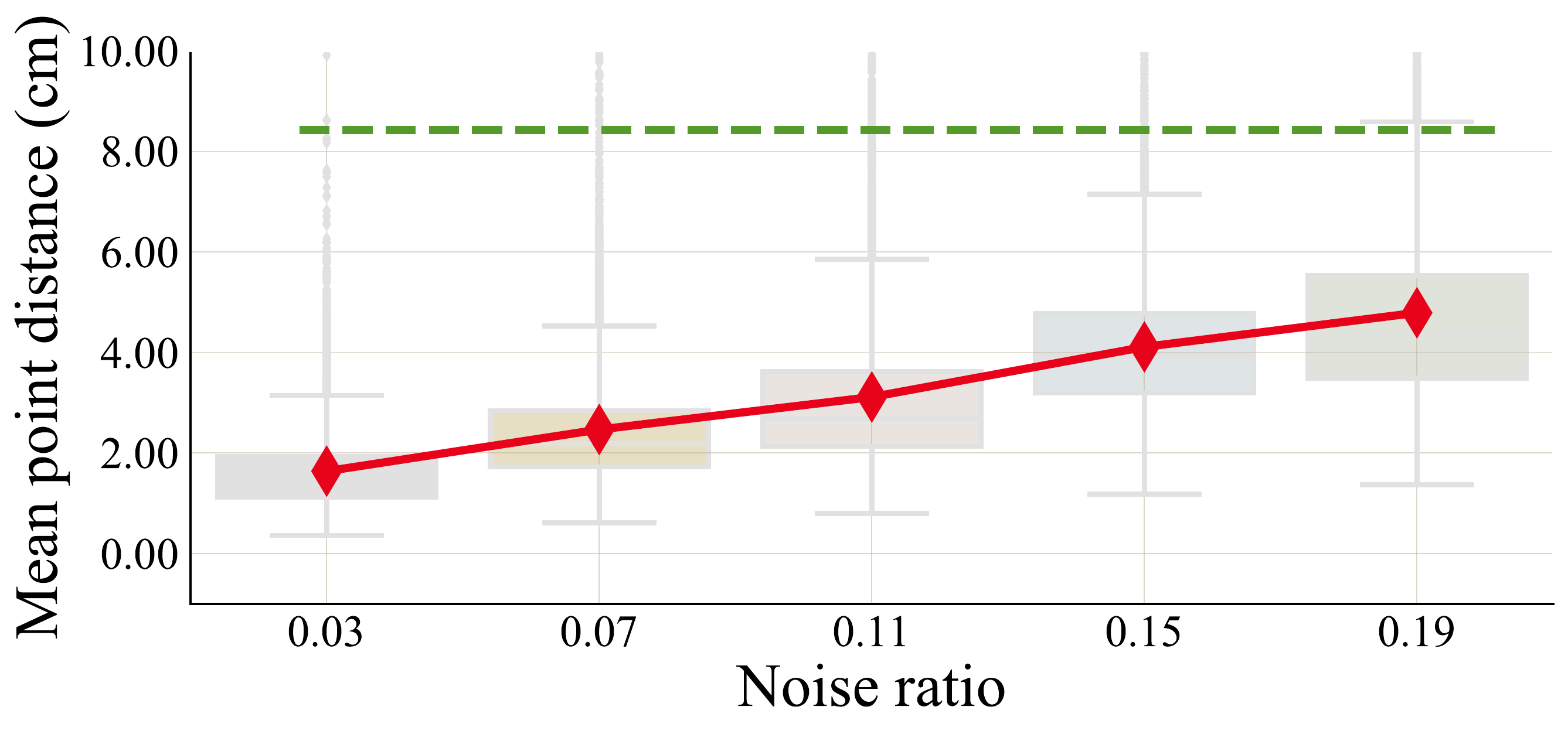}
 \caption{\nrsfm with noise perturbation. The red solid line is ours while the
  green dashed line is the lowest error achieved by baselines with \emph{no}
  noise perturbation.}
 \label{fig:noise}
\end{figure}

To analyze the robustness of our method, we
re-train the neural network for Subject 70 using projected points with
Gaussian noise perturbation. The results are summarized in Figure~\ref{fig:noise}.
The noise ratio is defined as $\Vert \text{noise} \Vert_F / \Vert \Wv \Vert_F$.
One can see that our method gets far more precise reconstructions even
when adding up to $20\%$ noise to our image coordinates compared to baselines
with no noise perturbation. This experiment clearly demonstrates the
robustness of our model and its high accuracy against state-of-the-art works.

\subsubsection{Missing data}
Landmarks are not always visible from the camera owing to the occlusion by other
objects or itself. In the present paper, we focus on a complete measurement
situation not accounting for invisible landmarks. However, thanks to recent
progress in deep-learning-based depth map
reconstruction from sparse observations~\cite{chen2018estimating,
 mal2018sparse, li2018depth, liao2017parse, cadena2016multi}, our central
pipeline of DNN can be easily adapted to handling missing data.

\subsection{S\textbf{\textit{f}}C on IKEA furnitures}
We now apply our method to the application of \sfc using IKEA dataset~\cite{lpt2013ikea,
 wu2016single}. The IKEA dataset contains four object categories: bed, chair,
sofa, and table. For each object category, we employ all annotated 2D point
clouds and augment them with 2K ones projected from the 3D ground-truth
using randomly generated orthonormal cameras\footnote{Augmentation is utilized
 due to limited valid frames, because the ground-truth cameras are
 partially missing.}. We compare our method against the
baselines~\cite{dai2014simple, kong2016sfc} again using the shape error ratio
metric. The error evaluated on real images are reported and summarized into
Table~\ref{tab:sfc}. One can observe that our method outperforms baselines
with a large margin, clearly showing the superiority of our model. Table~\ref{tab:sfc}
from another perspective reveals the dilemma suffered by baselines of
restricting ill-possedness and modeling high variance of object category. For
qualitative evaluation, we randomly select frames from each object category
and show them in Figure~\ref{fig:sfc}. It shows that our model successfully
learns the intra-category shape variation and reconstructed landmarks
effectively depict the 3D geometry of objects.

\begin{table}[t]
 \centering
 \footnotesize
 \begin{tabular}{C{1.2cm}C{1.2cm}C{1.2cm}C{1.2cm}C{1.2cm}}
  \specialrule{0.8pt}{1pt}{1pt}
                              & Bed             & Chair           & Sofa            & Table           \\
  \hline
  Simple~\cite{dai2014simple} & 17.81\%         & 33.32\%         & 14.78\%         & 12.40\%         \\
  SfC~\cite{kong2016sfc}      & 22.51\%         & 27.58\%         & 13.35\%         & 11.78\%         \\
  Ours                        & \textbf{0.23\%} & \textbf{1.15\%} & \textbf{0.35\%} & \textbf{0.81\%} \\
  \specialrule{0.8pt}{1pt}{1pt}
 \end{tabular}
 \caption{Quantitative comparison against state-of-the-art algorithms in \sfc
  task. Results are evaluated by shape error ratio. Our method outperforms
  others in all four object categories with a large margin.}
 \label{tab:sfc}
\end{table}

\subsection{Shape from single-view landmarks}
Even though almost all \nrsfm algorithms learn a shape dictionary from 2D
projections, none of them apply the learned dictionary to unseen data. This is
because all of them are facing the difficulty of handling large amount of
images and thus cannot generalize well. In this experiment, we show the
generalization of our learned dictionary by evaluating it using sequences invisible to training. Specifically,
we follow the same training and evaluation scheme in~\cite{zhou20153d},
training with Subject~86 in CMU MoCap and evaluating on Subject 13, 14 and 15.
We compare our model to methods for human pose
estimation~\cite{ramakrishna2012reconstructing, zhou20153d} following the same
error metrics in~\cite{zhou20153d}. It is worth mentioning that all baselines
learn shape dictionaries directly from 3D ground-truth, but our method learns
such dictionaries purely from 2D projections~(\ie no 3D supervision). Even in
such an unfair scenario, our method achieves competitive results as summarized
in Table~\ref{tab:recover}. This clearly demonstrates that our method
effectively learns the underlying geometry from pure 2D projections with no
need for 3D supervision, and the learned dictionaries generalize well to unseen
data.

\subsection{Coherence as guide}
\begin{figure}[H]
 \centering
 \includegraphics[width=\linewidth]{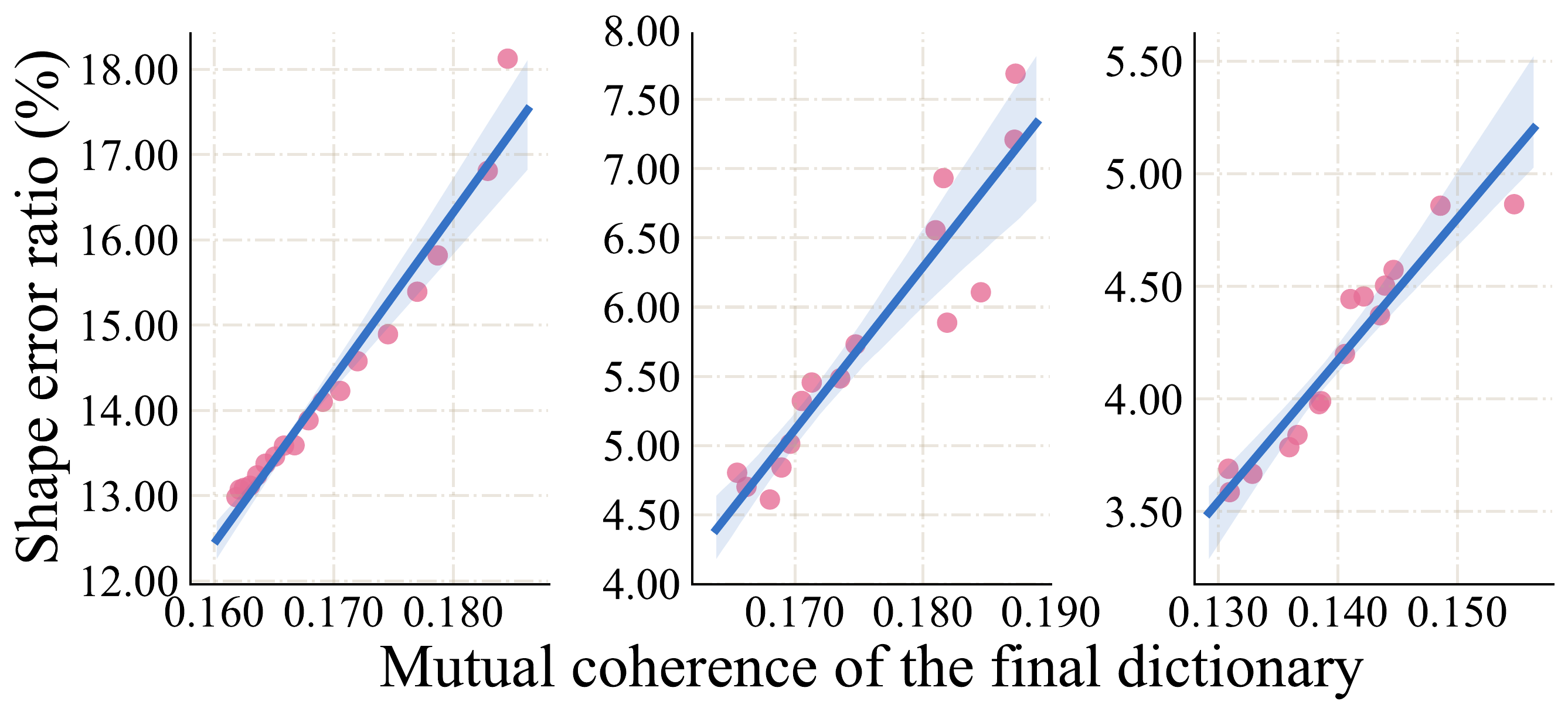}
 \caption{A scatter plot of the shape error ratio in percentage against the
  final dictionary coherence. A line is fitted based on the data. The left
  comes from subject 05, the middle from subject 18, the right from subject 64.}
 \label{fig:coherence}
\end{figure}

As explained in Section~\ref{sec:mlscm}, every sparse code~$\psiv_i$ is
constrained by its subsequent representation and thus the quality of code
recovery depends less on the quality of the corresponding dictionary. However,
this is not applicable to the final code~$\psiv_n$, making it least constrained
with the most dependency on the final dictionary~$\Dv_n$. From this
perspective, the quality of the final dictionary measured by mutual
coherence~\cite{donoho2006stable} could serve as a lower bound of the entire
system. To verify this, we compute the error and coherence in a fixed interval
during training in \nrsfm experiments. We consistently observe strong
correlations between 3D reconstruction error and the mutual coherence of the
final dictionary. We plot this relationship in Figure~\ref{fig:coherence}. We
thus propose to use the coherence of the final dictionary as a measure of model
quality for guiding training to efficiently avoid over-fitting especially when
3D evaluation is not available. This improves the utility of our deep \nrsfm in
future applications without 3D ground-truth.

\begin{table}[t]
 \centering
 \footnotesize
 \begin{tabular}{C{1.2cm}C{1.2cm}C{1.2cm}C{1.2cm}C{1.2cm}}
  \specialrule{0.8pt}{1pt}{1pt}
             & PMP   & Alternate & Convex         & Ours           \\
  \hline
  Subject 13 & 0.390 & 0.293     & 0.259          & \textbf{0.229} \\
  Subject 14 & 0.393 & 0.308     & \textbf{0.258} & 0.261          \\
  Subject 15 & 0.340 & 0.286     & 0.204          & \textbf{0.200} \\
  \specialrule{0.8pt}{1pt}{1pt}
 \end{tabular}
 \caption{Comparison of our method against the state-of-the-art algorithms in
  single image human pose estimation task. Our method achieves competitive
  results using solely 2D projections while all others learn from
  3D ground truth.}
 \label{tab:recover}
\end{table}

\section{Conclusion}
In this paper, we proposed multi-layer sparse coding as a novel prior
assumption for representing 3D non-rigid shapes and designed an innovative
encoder-decoder neural network to solve the problem of \nrsfm using no 3D
supervision. The proposed DNN was derived by generalizing the classical sparse
coding algorithm ISTA to a block sparse scenario. The proposed DNN architecture
is mathematically interpretable as a \nrsfm multi-layer sparse dictionary
learning problem. Extensive experiments demonstrated our superior performance
against the state-of-the-art methods and the impressive generalization to
unseen data. Finally, we propose to use the coherence of the final dictionary
as a generalization measure, offering a practical way to avoid over-fitting and
selecting the best model without 3D ground-truth.

{\small
\bibliographystyle{ieee}
\bibliography{egbib}
}

\end{document}